# Towards Effective Sentence Simplification for Automatic Processing of Biomedical Text

**Siddhartha Jonnalagadda**[*], **Luis Tari**[**], **Jörg Hakenberg**[**], **Chitta Baral**[**], **Graciela Gonzalez**[*]
[*]Department of Biomedical Informatics, Arizona State University, Phoenix, AZ 85004, USA.
[**]Department of Computer Science and Engineering, Arizona State University, Tempe, AZ 85281, USA.
Corresponding author: ggonzalez@asu.edu

## Abstract

The complexity of sentences characteristic to biomedical articles poses a challenge to natural language parsers, which are typically trained on large-scale corpora of non-technical text. We propose a text simplification process, bioSimplify, that seeks to reduce the complexity of sentences in biomedical abstracts in order to improve the performance of syntactic parsers on the processed sentences. Syntactic parsing is typically one of the first steps in a text mining pipeline. Thus, any improvement in performance would have a ripple effect over all processing steps. We evaluated our method using a corpus of biomedical sentences annotated with syntactic links. Our empirical results show an improvement of 2.90% for the Charniak-McClosky parser and of 4.23% for the Link Grammar parser when processing simplified sentences rather than the original sentences in the corpus.

## 1 Introduction

It is typical that applications for biomedical text involve the use of natural language syntactic parsers as one of the first steps in processing. Thus, the performance of the system as a whole is largely dependent on how well the natural language syntactic parsers perform.

One of the challenges in parsing biomedical text is that it is significantly more complex than articles in typical English text. Different analysis show other problematic characteristics, including inconsistent use of nouns and partial words (Tateisi & Tsujii, 2004), higher perplexity measures (Elhadad, 2006), greater lexical density, plus increased number of relative clauses and prepositional phrases (Gemoets, 2004), all of which correlate with diminished comprehension and higher text difficulty. These characteristics also lead to performance problems in terms of computation time and accuracy for parsers that are trained on common English text corpus.

We identified three categories of sentences: 1) normal English sentences, like in Newswire text, 2) normal biomedical English sentences – those sentences which can be parsed without a problem by Link Grammar-, and 3) complex biomedical English sentences – those sentences which can't be parsed by Link Grammar. Aside from the known characteristics mentioned before, sentences in the third group tended to be longer (18% of them had more than 50 words, while only 8% of those in group 2 and 2% of those in group 1 did). It has been observed that parsers perform well with sentences of reduced length (Chandrasekar & Srinivas, 1997; Siddharthan, 2006).

In this paper, we explore the use of text simplification as a preprocessing step for general parsing to reduce length and complexity of biomedical sentences in order to enhance the performance of the parsers.

## 2 Methods

There are currently many publicly available corpora of biomedical texts, the most popular among them being BioInfer, Genia, AImed, HPRD 50, IEPA, LLL and BioCreative1-PPI. Among these corpora, BioInfer includes the most comprehensive collection of sentences and careful annotation for links of natural parser, in both the Stanford and Link Grammar schemes. Therefore, we chose the BioInfer corpus, version 1.1.0 (Pyysalo et al., 2007), containing 1100 sentences for evaluating the effectiveness of our simplification method on the performance of syntactic parsers. The method includes syntactic and non-syntactic transformations, detailed next.

## 2.1 Non-syntactic transformation

We group here three steps of our approach: 1. pre-processing through removal of spurious phrases; 2. replacement of gene names; 3. replacement of noun phrases.

To improve the correctness of the parsing, each biomedical sentence is first preprocessed to remove phrases that are not essential to the sentence. This includes removal of *section indicators*, which are phrases that specify the name of the section at the beginning of the sentence, plus the removal of phrases in parentheses (such as citations and numbering in lists). Also, partially hyphenated words are transformed by combining with the nearest word that follows or precedes the partial hyphenated word to make a meaningful word. For instance, the phrase "alpha- and beta-catenin" is transformed into "alpha-catenin and beta-catenin".

Occurrences of multi-word technical terms and entity names involved in biomedical processes are common in biomedical text. Such terms are not likely to appear in the dictionary of a parser (perplexity is high), and will force it to use morpho-guessing and unknown word guessing. This is time consuming and prone to error. Thus, unlike typical text simplification that emphasizes syntactic transformation of sentences, our approach utilizes a named entity recognition engine, BANNER (Leaman & Gonzalez, 2008), to replace multi-word gene names with single-word placeholders.

Replacement of gene names with single elements is not enough, however, and grammatical category (i.e. singular or plural) of the element has to be considered. Lingpipe (Alias-i, 2006), a shallow parser for biomedical text, identifies noun phrases and replaces them with single elements. A single element is considered singular when the following verb indicates a third-person singular verb or the determiner preceded by the element is either "a" or "an". Otherwise it is considered as plural and an "s" is attached to the end of the element.

## 2.2 Syntactic transformation

The problem of simplifying long sentences in common English text has been studied before, notably by Chandrasekar & Srinivas (1997) and Siddharthan (2006). However, the techniques used in these studies might not totally solve the issue of parsing biomedical sentences. For example, using Siddharthan's approach, the biological finding "The Huntington's disease protein interacts with p53 and CREB-binding protein and represses transcription", and assuming multi-word nouns such as "CREB-binding protein" do not present a problem, would be simplified to:

> "The Huntington's disease protein interacts with p53. The Huntington's disease protein interacts with CREB-binding protein. The Huntington's disease protein represses transcription."

Our method transforms it to "GENE1 interacts with GENE2 and GENE3 and represses transcription." Both decrease the average sentence length, but the earlier distorts the biological meaning (since the Huntington's disease protein might not repress transcription on its own), while the latter signifies it.

While replacement of gene names and noun phrases can reduce the sentence length, there are cases when the sentences are still too complex to be parsed efficiently. We developed a simple algorithm that utilizes *linkages* (specific grammatical relationships between pairs of words in a sentence) of the Link Grammar parser (Sleator, 1998) and punctuations for splitting sentences into clauses. An example in Figure 1 illustrates the main part of the algorithm. Each linkage has a primary *link type* in CAPITAL followed by secondary link type in short. The intuition behind the algorithm is to try to identify independent clauses from complex sentences. The first step is to split the sentence *ST* into clauses $c_1$, $c_2$, $c_3$ and $c_4$ based on commas. $c_1$ is parsed using the Link Grammar parser, but $c_1$ cannot be a sentence as there is no "S" link in the linkage of $c_1$. $c_2$ is then attached to $c_1$ and the linkage of "$c_1$, $c_2$" does not contain a "S" link as well. "$c_1$, $c_2$, $c_3$." is recognized as a sentence, since the linkage contains an "S" link, indicating that it is a sentence, as well as the linkage of $c_4$. So the algorithm returns two sentences $ST_1$ and $ST_2$ for *ST*.

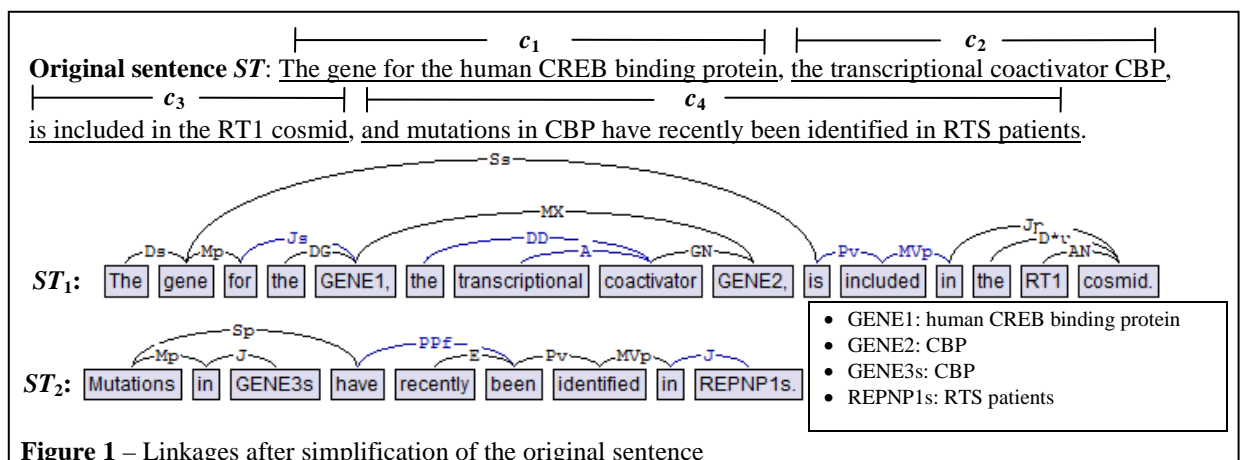

**Figure 1** – Linkages after simplification of the original sentence

## 3 Results

Our method has the greatest impact on the performance of Link Grammar (LG), which lies at the core of BioLG (Pyysalo et al., 2006). However, it also has a significant impact on the self-training biomedical parser by McClosky & Charniak (CM), which is currently the best parser available for biomedical text.

**3.1 Rudimentary statistics of the results of simplification:** After the simplification algorithm was tested on the 1100 annotated sentences of the BioInfer corpus, there were 1159 simplified sentences because of syntactic transformation (section 2.2). The number of words per sentence showed a sharp drop of 20.4% from 27.0 to 21.5. The Flesh-Kincaid score for readability dropped from 17.4 to 14.2. The Gunning Fog index also dropped by 18.3% from 19.7 to 16.1.

| Pre-processing | Replacement of gene names | Replacement of noun phrases | Syntactic Simplification |
|---|---|---|---|
| 359 | 1082 | 915 | 91 |

Table 1: Sentences processed in each stage

**3.2 Impact of simplification on the efficiency of parsing:** We inputted the BioInfer corpus to LG and CM. If LG cannot find a complete linkage, it invokes its *panic mode*, where sentences are returned with considerably low accuracy. Out of the 1100 original sentences in the corpus, 219 went into panic mode. After processing, only 39 out of 1159 simplified sentences triggered panic mode (a 16.4% improvement in efficiency). The average time for parsing a sentence also dropped from 7.36 secs to 1.70 secs after simplification.

**3.3 Impact of simplification on the accuracy of parsing:** Let Σg, Σo and Σs, respectively be the sets containing the links of the gold standard, the output generated by the parser on original sentences and the output generated by the parser on simplified sentences. We denote a link of type Π between the tokens $\Phi_1$ and $\Phi_2$ by $(\Pi,\Phi_1,\Phi_2)$. In the case of the original sentences, the tokens $\Phi_1$ and $\Phi_2$ are single-worded. So, $(\Pi,\Phi_1,\Phi_2)$ is a true positive iff $(\Pi,\Phi_1,\Phi_2)$ belongs to both Σg and Σo, false positive iff it only belongs to Σo and false negative iff it only belongs to Σg. In the case of simplified sentences, the tokens $\Phi_1$ and $\Phi_2$ can have multiple words. So, $(\Pi,\Phi_1,\Phi_2)$ which belongs to Σs is a true positive iff $(\Pi,\Phi'_1,\Phi'_2)$ belongs to Σg where $\Phi'_1$ and $\Phi'_2$ are respectively one of the words in $\Phi_1$ and $\Phi_2$. Additionally, $(\Pi,\Phi_1,\Phi_2)$ which belongs to Σg is not a false negative if $\Phi_1$ and $\Phi_2$ are parts of a single token of a simplified sentence. For measuring the performance of a parser, the nature of linkage is most relevant in the context of the sentence in consideration. So, we calculate precision and recall for each sentence and average them over all sentences to get the respective precision and recall for the collection.

|  | Precision | Recall | f-measure |
|---|---|---|---|
| CM | 77.94% | 74.08% | 75.96% |
| BioSimplify + CM | 82.51% | 75.51% | 78.86% |
| **Improvement** | **4.57%** | **1.43%** | **2.90%** |
|  |  |  |  |
| LG | 72.36% | 71.65% | 72.00% |
| BioSimplify + LG | 78.30% | 74.27% | 76.23% |
| **Improvement** | **5.94%** | **2.62%** | **4.23%** |

Table 2: Accuracy of McClosky & Charniak (CM) and Link Grammar (LG) parsers based on Stanford dependencies, with and without simplified sentences.

In order to compare the effect of BioSimplify on the two parsers, a converter from Link Grammar to Stanford scheme was used (Pyysalo et al, 2007: precision and recall of 98% and 96%). Results of this comparision are shown in Table 2. On CM and LG, we were able to achieve a considerable improvement in the f-measures by 2.90% and 4.23% respectively. CM demonstrated an absolute error reduction of 4.1% over its previous best on a different test set. Overall, bioSimplify leverages parsing of biomedical sentences, increasing both the efficiency and accuracy.

## 4 Related work

During the creation of BioInfer, noun phrase macro-dependencies were determined using a simple rule set without parsing. Some of the problems related to parsing noun phrases were removed by reducing the number of words by more than 20%. BioLG enhances LG by expansion of lexicons and the addition of morphological rules for biomedical domain. Our work differs from BioLG not only in utilizing a gene name recognizer, a specialized shallow parser and syntactic transformation, but also in creating a preprocessor that can improve the performance of any parser on biomedical text.

The idea of improving the performance of deep parsers through the integration of shallow and deep parsers has been reported in (Crysmann et al., 2002; Daum et al., 2003; Frank et al., 2003) for non-biomedical text. In BioNLP, extraction systems (Jang et al., 2006; Yakushiji et al., 2001) used shallow parsers to enhance the performance of deep parsers. However, there is a lack of evaluation of the correctness of the dependency parses, which is crucial to the correctness of the extracted systems. We not only evaluate the correctness of the links, but also go beyond the problem of relationship extraction and empower future researchers in leveraging their parsers (and other extraction systems) to get better results.

## 5 Conclusion and Future work

We achieved an f-measure of 78.86% using CM on BioInfer Corpus which is a 2.90% absolute reduction in error. We achieved a 4.23% absolute reduction in error using LG. According to the measures described in section 3.1, the simplified sentences of BioInfer outperform the original ones by more than 18%. Our method can also be used with other parsers. As future work, we will demonstrate the impact of our simplification method on other text mining tasks, such as relationship extraction.

## Acknowledgments

We thank Science Foundation Arizona (award CAA 0277-08 Gonzalez) for partly supporting this research. SJ also thanks Bob Leaman and Anoop Grewal for their guidance.